\documentclass[letterpaper, 10 pt, conference]{ieeeconf}  

\usepackage{url}
\usepackage{graphicx}
\usepackage[ruled,vlined]{algorithm2e}
\SetKwIF{If}{ElseIf}{Else}{if}{}{else if}{else}{end if}%
\SetKwFor{While}{while}{}{end while}%
\SetKwFor{For}{for}{}{end for}%
\usepackage{caption}
\usepackage{color}
\usepackage{soul}
\usepackage{subcaption}
\usepackage{float}
\usepackage{stfloats}
\IEEEoverridecommandlockouts                              

\title{\LARGE \bf
Robotic Barrier Construction through Weaved, Inflatable Tubes
}

\author{Heather Jin Hee Kim$^{1}$*, Haron Abdel-Raziq$^{2}$*, Xinyu Liu$^{3}$, Alexandra Young Siskovic$^{3}$, Shreyas Patil$^{1}$, \\ Kirstin H. Petersen$^{2}$, and Hsin-Liu (Cindy) Kao$^{1}$
\thanks{This work was funded by a Packard Fellowship for Science and Engineering and NSF grants CNS-2042411 and IIS-2047249.}
\thanks{$^{1}$Department of Human-Centered Design,
        $^{2}$Department of Electrical and Computer Engineering,
        and $^{3}$Sibley School of Mechanical Engineering, Cornell University, Ithaca NY-14853, USA.
        {\tt\small Emails: [jk2768, hma49, xl598, ays28, sp2544, kirstin, cindykao] @cornell.edu}}%
\thanks{$^{*}$These authors contributed equally to this work.}
}

\begin{document}

\maketitle
\thispagestyle{empty}
\pagestyle{empty}

\begin{abstract}
In this article, we present a mechanism and related path planning algorithm to construct light-duty barriers out of extruded, inflated tubes weaved around existing environmental features. Our extruded tubes are based on everted vine-robots and in this context, we present a new method to steer their growth. We characterize the mechanism in terms of accuracy resilience, and, towards their use as barriers, the ability of the tubes to withstand distributed loads. We further explore an algorithm which, given a feature map and the size and direction of the external load, can determine where and how to extrude the barrier. 
Finally, we showcase the potential of this method in an autonomously extruded two-layer wall weaved around three pipes. While preliminary, our work indicates that this method has the potential for barrier construction in cluttered environments, e.g. shelters against wind or snow. Future work may show how to achieve tighter weaves, how to leverage weave friction for improved strength, how to assess barrier performance for feedback control, and how to operate the extrusion mechanism off of a mobile robot.
\end{abstract}

\section{INTRODUCTION}

The introduction of robot automation to the construction sector is promising for both traditional~\cite{melenbrink2020site} and more futuristic~\cite{petersen2019review} tasks, and spans from construction of human habitats~\cite{aejmelaeus2020rock,werfel2014designing} and extra-terrestrial support structures~\cite{saboia2019autonomous,benvenuti2013living}, to environmental protection barriers~\cite{liu2021planning}. In most tasks, a key consideration involves whether to use in-situ or brought material; in-situ materials such as sand, debris, and rocks may be hard to manipulate and/or modify, but are essentially free and unlimited, whereas brought material can be customized for the robot and building application, but requires transport to the construction site. 

\begin{figure}[b]
    \centering
    \includegraphics[width =0.4\textwidth]{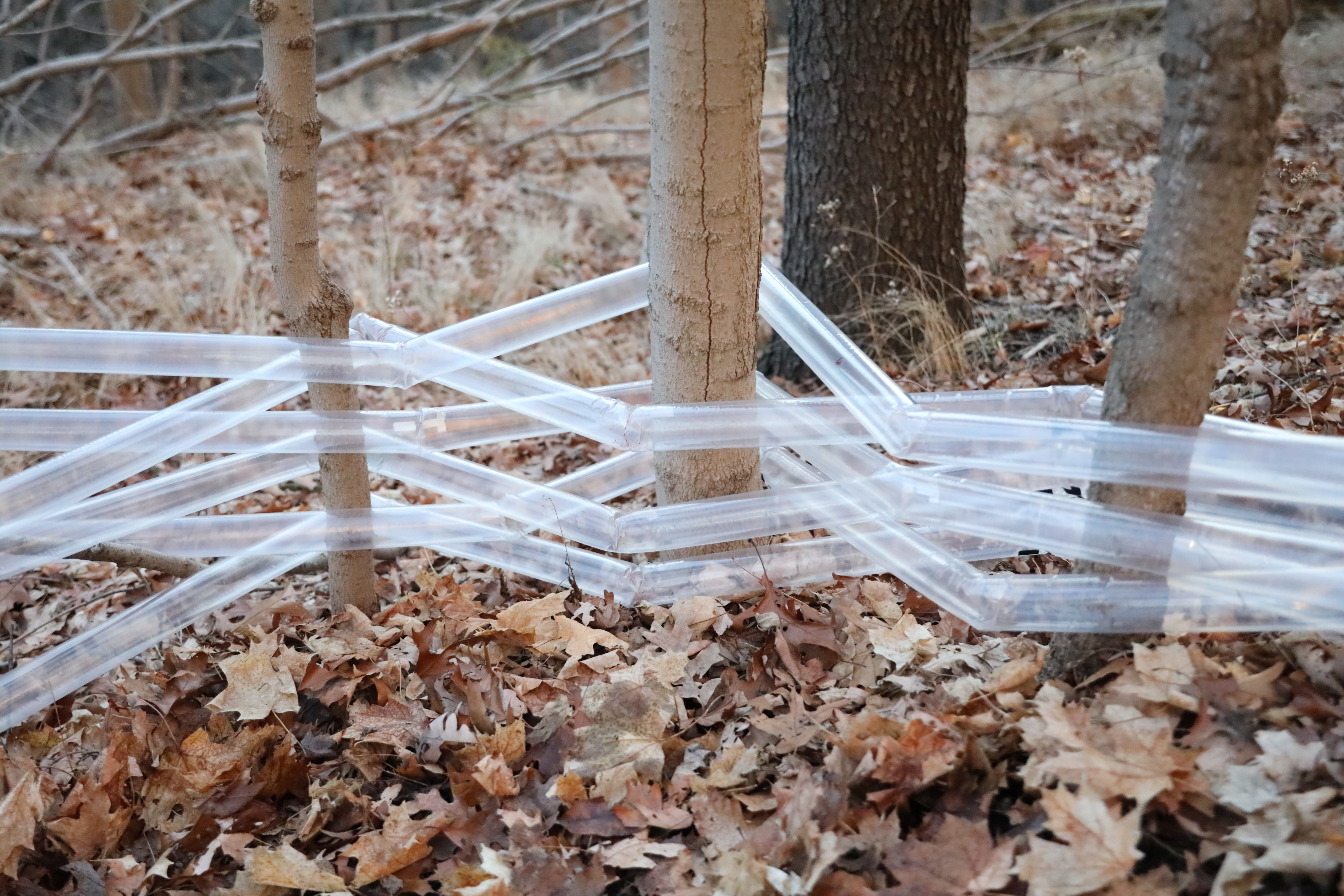}
    \caption{Example shelter wall composed of inflated tubes interleaved between existing environmental features.}
    \label{fig:overview}
\end{figure}

In this paper, we investigate the use of extruded tubes weaved around existing environmental features to quickly assemble light-duty barriers, e.g. against wind or snow (Fig.~\ref{fig:overview}). By extruded tubes, we are referring to the recent advancement in `vine-robots' where everted, flexible polymer tubes can be inflated several meters out from their deployment mechanism~\cite{hawkes2017soft}. Our proposed method is promising for a number of reasons: 1) The use of inflated tubes means that the robot payload is small compared to the size of the inflated structures. 2) The use of existing environmental features, such as tree trunks, adds structural stability. 3) Similar to the erection of shelter tarps, grown tubes are inexpensive and leave only a small amount of waste, yet are much easier for autonomous robots to deploy because they do not require cooperative manipulation to stretch material between distanced geometric features. 4) Finally, because the robot does not have to maneuver across the construction site the demands on mobility are lower (imagine deploying these structures across a steep forested slope).

Specifically, we investigate the potential of this technology by introducing and characterizing a novel mechanism to steer tube extrusion (Sec.~\ref{sec:mec}); we investigate the tube behavior under distributed loads (Sec.~\ref{sec:char}); we propose a suitable path planning algorithm for the weaved barrier, assuming full prior knowledge of the environment and the external load (Sec.~\ref{sec:alg}); and we demonstrate an autonomously extruded two-layer structure weaved around three pipes and explore its resilience to imperfections in the map (Sec.~\ref{sec:demo}). While our findings are preliminary, our work suggests that extruded, inflated barriers can accommodate an important niche of applications in autonomous robotic construction (Sec.~\ref{sec:conclusion}). 

\section{WEAVING MECHANISM}
\label{sec:mec}

Everted vine-robots were first introduced in 2017~\cite{hawkes2017soft}, and have been shown in a wealth of applications and permutations since, including wearable haptic devices; smart hospital beds; deployable antennas; and search and navigation in constrained environments such as archaeological digs, underwater ecosystems, in granular materials, and inside the human body~\cite{blumenschein2020design}. In the following sub-sections we briefly describe how we implemented the extrusion mechanism, and then detail and argue for a new vine-robot steering mechanism. However, it is worth noting that our approach to barrier construction may also work for other vine-robot steering mechanisms and other tube-inflation robots~\cite{satake2020novel}. 

\subsection{Growth Mechanism}

\begin{figure}[t]
    \centering
    \includegraphics[width =0.5\textwidth]{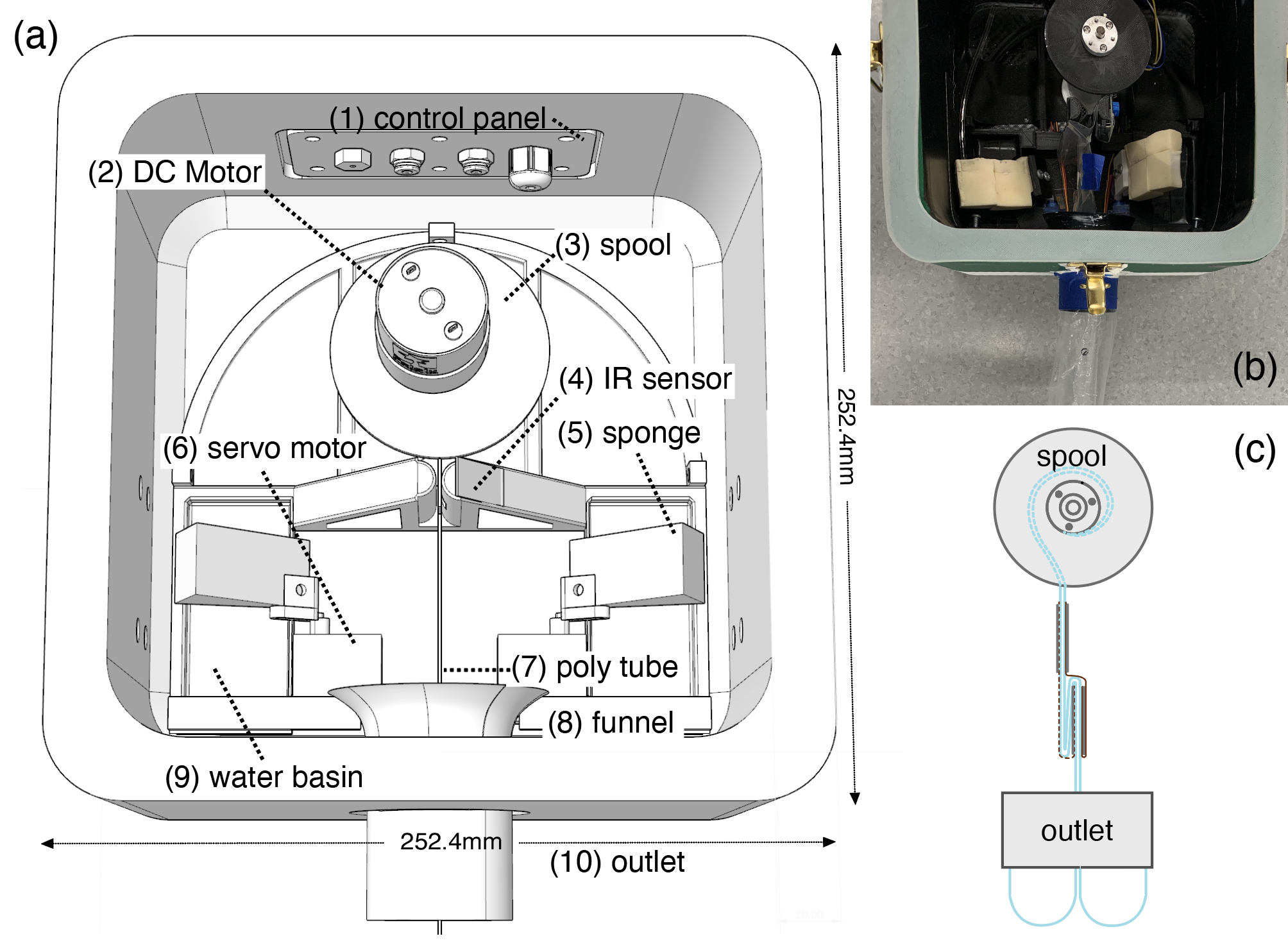}
    \caption{An exploded view of the extrusion mechanism used in this work is shown in (a). Manufactured chassis and components are photographed in (b). A simplified schematic of the reeled tube and fold is shown in (c).}
    \label{fig:grow}
\end{figure}

We replicated the everted vine-robot originally presented in~\cite{hawkes2017soft} with the design shown in Fig.~\ref{fig:grow}. We manufactured the (rigid) chassis using PLA filament on a low-end 3D printer (Lulzbot TAZ 6 with a 0.8mm nozzle). The chassis was composed of a simple cube box with a lid, of side length $\sim$250mm and 20mm thick walls to withstand high internal pressure. 
We took several measures to secure an airtight seal: 1) we configured the printer to add 10 perimeter layers and heat treated the surface of the printed parts to avoid gaps between extruded layers; 2) we applied caulk around all screw holes to keep the positive pressure contained to the internal chamber; and 3) we secured the lid using four draw latches and a 5mm thick rim of Ecoflex 00-30 polymer.

Inside the chassis we mounted a geared DC motor (Polulu item number 4744), and attached its output shaft to a 3D printed spool using a 2-sided coupler and a non-flanged ball bearing from Servocity (1600 Series). The flexible tubing was wrapped around this spool; we used $t = $ 2mil thick low-density polyethylene (LDPE) tubing with a $D_{flat} =$ 3" lay-flat diameter from ULine. The tube extruded out through a 50mm hole with a 3D printed funnel to avoid snagging. For debugging ease, we mounted all control circuitry on the outside of the chassis. This circuitry included an Arduino Uno and a one-directional motor driver composed of an N-channel MOSFET and some resistors. Finally, we inflated the robot mechanism using a Campbell Hausfeld FP209402 compressor. 

\subsection{Existing Steering Mechanisms}
\label{sec:old_steering}

To weave the tube back and forth between existing environmental features, such as the tree trunks in Fig.~\ref{fig:overview}, we must be able to steer the direction of extrusion in 2D. More specifically, given the need for a barrier, we assume that each inflated tube will consist predominantly of long straight segments, interspersed by a brief series of discrete bends large enough to direct the extrusion in the opposite direction around the following feature. We also assume that the barrier material is expendable, i.e. that there is no need to retract and/or reconfigure the tube once in place.

Many steering mechanisms for everting vine-robots have been demonstrated.
Passive mechanisms include tube pre-configuration~\cite{hawkes2017soft} or tubes that leverage collisions with the environment~\cite{greer2020robust}. The former method is useful only if the environment is known before robot deployment; the latter would severely limit the subset of feature-pairs the tubes could be weaved around. 
Active steering mechanisms are numerous, and most are based on active compression. Pneumatic artificial muscles or tendons can be mounted along the tube to produce strain-induced curvatures ~\cite{greer2017series,blumenschein2018tip}. However, since the same strain is applied all along the tube, these methods do not lend themselves well to the alternating bend directions which are needed for weaves. Adding independently controlled segments along the tube would increase control and design complexity and does not scale well with length.  
Leveraging active extension, tip-localized steering involves pre-loaded strain that can be released only at the tip of the tube, implemented either through air pockets driven by air channels that run all along the length of tube~\cite{hawkes2017soft} or through a servo-based mechanism that sits at the tip of the tube able to cut pre-tensioned cables~\cite{cinquemani2020design}. While this tip-localized steering would enable alternating curvatures, the air channels would add extrusion friction and the servo-based mechanism mounted at the tip could add the risk of snagging. 

\subsection{Steering through Active Extension}
\label{sec:new_steering}

Motivated by the need for weaved structures, we designed a new steering mechanism based on extension, rather than compression, where the active steering mechanism is located at the base of the robot. 

\subsubsection{Concept}

\begin{figure}[t]
    \centering
    \includegraphics[width =0.45\textwidth]{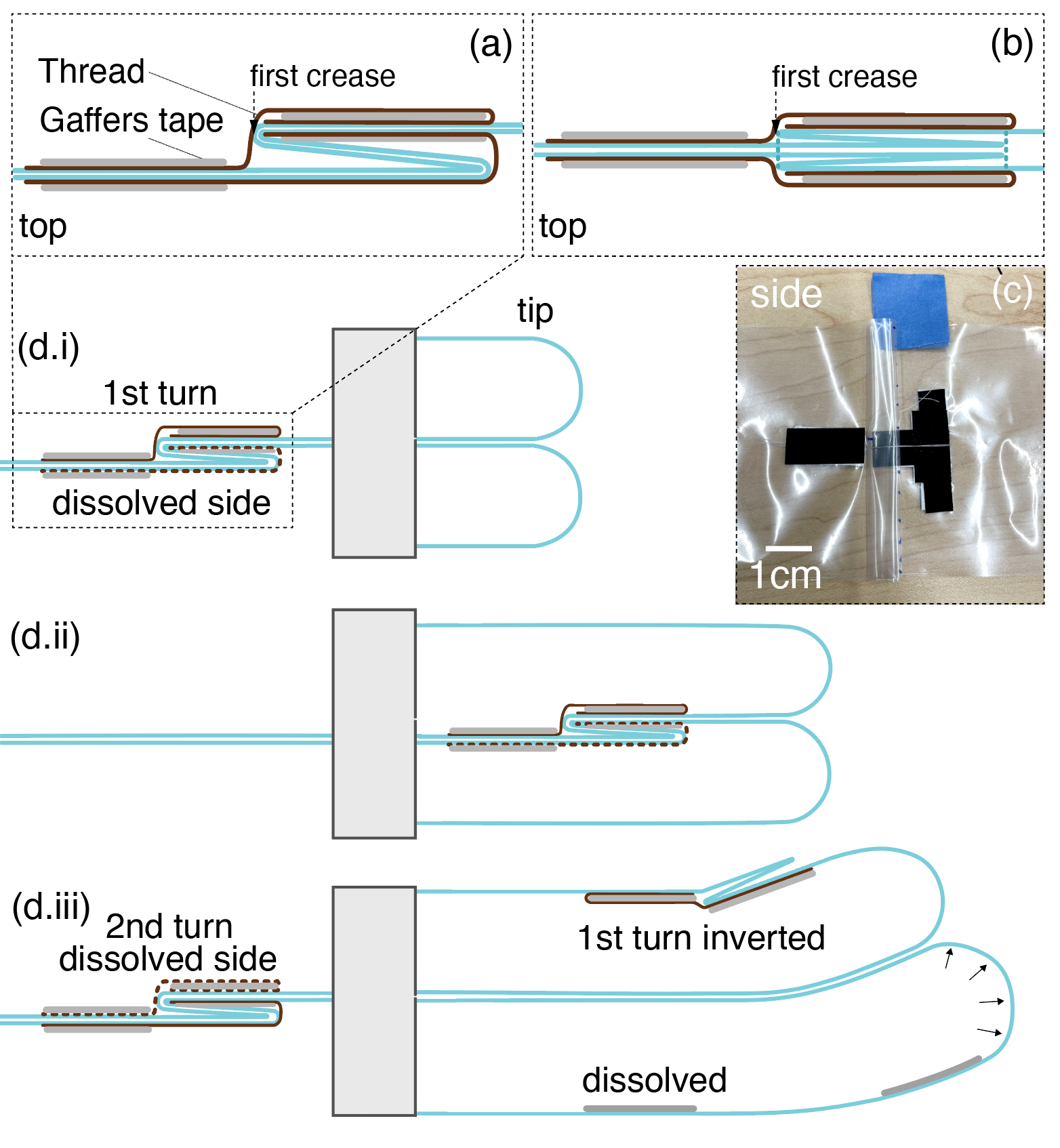}
    \caption{Steering mechanism based on an active extension of the tube. (a-b) The tube is pre-folded at regular intervals using either method (a) or (b). Folds are held in place by a thread on each side of the tube. (c) Example of a real fold. (d) Concept animation. If one side of the fold is released, the tube will bend in the direction of the remaining thread, when pressurized and extruded.}
    \label{fig:steering}
\end{figure}

In this mechanism the tubes are folded double at regular intervals. The folds are held in place by a thread (Fig.~\ref{fig:steering}), and these threads can be released at the base of the robot in anticipation of a desired turn. When the affected part of the tube reaches the tip and becomes pressurized, it will bend towards the remaining thread unfolding the side of the tube where the thread was released. This method is very simple to implement and the design complexity remains constant independent of the number of turns needed in the inflated structure. When the tube is extruded straight, this simplicity comes at the cost of added material (folds can be left in place) or extrusion time (both threads holding a fold in place can be released to eliminate the fold entirely). While it is not an issue for our application, it is also worth noticing that the method can only produce discrete turns and the created bends cannot be reversed. 

\subsubsection{Release mechanism}

The thread may be released in a number of ways. Arguably, the most intuitive would be a mechanical cutter; however, due to existing lab know-how, we decided to use water-soluble thread. We adhered the thread across the folds using Gaffers tape. To selectively dissolve the thread we mounted a servo (Miuzei 9G Micro Servo) with a sponge and a water basin in the chassis on either side of the tube. We used an infrared break-beam sensor (TCRT5000) to inform when the fold was in position and then selectively dissolved the thread on the side we wanted the tube to bend away from (Fig.~\ref{fig:steering}(d)). Repeated testing revealed that Vanish-Lite water soluble thread (diameter 0.09mm) works well, given an extrusion pressure of 5-8 PSI. With this thread, the water-filled sponges would dissolve the thread in 40s (practically, we found the most reliable method was to re-wet the sponge every 10s). 

\subsubsection{Folds}
We experimented with two types of folds shown in Fig.~\ref{fig:steering}(a-b, and c). In the first, the entire tube is folded over; in the second, the tube is folded symmetrically about itself.
We found empirically that the former worked more consistently upon extrusion, presumably because it poses less friction within the tube as it everts and turns. Here, the tube walls bifurcate easily until the first crease, where the folds snap off each other and immediately unfold on the side which was released. In fold type (b), the fold faces have to roll in conjunction with the tube walls, and the released side has to unfold subsequently causing greater stiffness in the mechanism overall. 

The fold length, $L_{fold}$, determines how far the tube will bend when it is pressurized and one side is released (Fig.~\ref{fig:fold-v-angle}(a-b)). Given the thread length, $L_{thread} \approx L_{fold}$, and the inflated diameter of the tube $D_{infl} = 2D_{flat}/\pi$, we can estimate the fold angle $\theta$ in radians given the following set of equations:

\begin{equation}
    \label{eq1}
    \theta = \frac{2L_{fold}}{D_{infl}+x}
\end{equation}
\begin{equation}
    \label{eq2}
    \theta = 2sin^{-1}(\frac{L_{fold}}{2x})
\end{equation}

\subsubsection{Repeatability} 
\begin{figure}[h]
    \centering
    \includegraphics[width =0.48\textwidth]{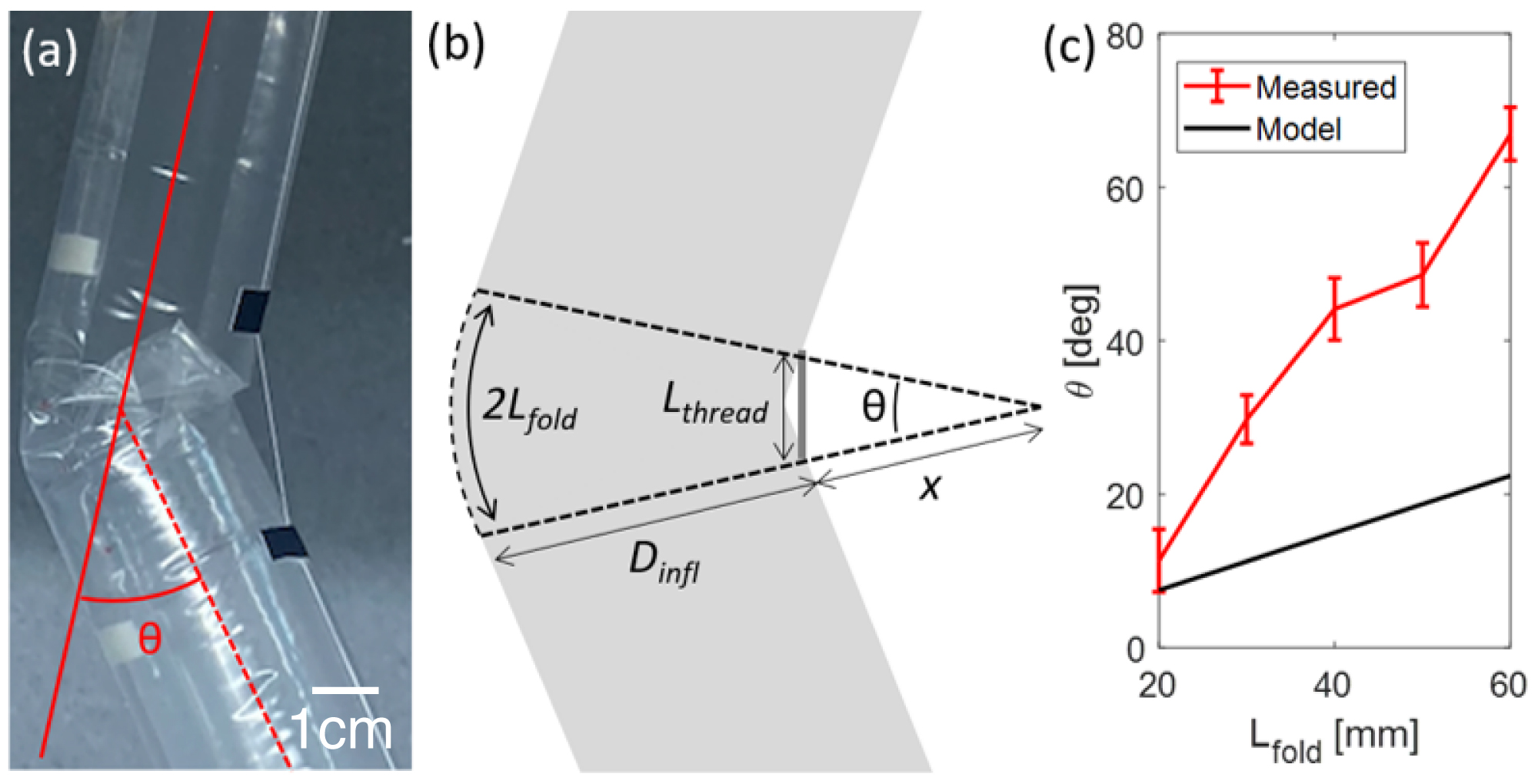}
    \caption{(a) Example fold. (b) Simplified fold model. (c) Measured and modeled correlation between fold length $L_{fold}$ and angle $\theta$. Measured data indicates mean and standard deviation over 5 successful trials.}
    \label{fig:fold-v-angle}
\end{figure}

We tested the repeatability of our mechanism by manually executing the release mechanism described above. The data is shown in Fig.~\ref{fig:fold-v-angle}(c). We found that the bend angles were reproducible and that $\theta$ is roughly proportional to $L_{fold}$ as expected. The exact parameters, however, do not agree with Eq.~\ref{eq1}-\ref{eq2}, presumably because of the messy fold that occurs on the inside of the bend. Better agreement with the model might be produced if the thread length, $L_{thread}$, was decreased.

Each data point in the graph shows the mean and standard deviation of 5 successful releases. With low fold lengths ($L_{fold}=$20mm), we recorded 5/7 successful trials. Failures occurred because the Gaffers tape would slip off the tube. With large fold lengths ($L_{fold}=$[50, 60]mm) we recorded 5/8 and 5/10 successful trials respectively. Here, failures would occur because the internal tube pressure was not sufficient to overcome the friction of the fold. This would cause a pressure build-up that would make the Gaffers tape slip off the tube again. At fold lengths of 40 and 50mm we recorded no failures. To mitigate these failures, stronger adhesion, e.g. based on epoxy, would be necessary. 

\section{Inflated Tubes as Barriers}
\label{sec:char}

Depending on the use case, there are many ways to evaluate the performance of light-duty barriers and shelter walls. At a high level, these include material usage/waste, cost, durability, strength, and permeability. 

As already discussed, inflated tubing inflicts minimal waste, similar in scale to erecting a tarp. For example, we can estimate the weight and packed volume of tubing needed for a 1$\times$1m$^2$ wall as follows. It would require ($N = 1m/D_{infl} = $) 21 stacked tubes. 
Not inflated, these 21 stacked tubes of $L_{tube} = 1m$ length would take up a volume of $V_{1\times1m^2} = 2 N t D_{flat} L_{tube} = 162.58cm^3$ or a cube with 5.5cm sides (the factor 2 stems from the fact that the tube has two walls). Given the density of LDPE, $\rho_{LDPE} = 0.91 g/cm^3$, this would amount to a payload of just 147.9g. 

Polyethylene tubing is generally used in packaging and medical applications, and is therefore easily obtainable and very low cost; we acquired 3000 feet for 66USD, i.e. 2.2 cents per foot of material. In terms of durability, having been developed for everything from electrical insulation to shopping bags, LDPE is resistant to impacts, moisture, radiation, temperature variations, and a wide range of chemicals. 

Finally, the ability of the tubes to withstand loads from the environment, e.g., the ability to withstand distributed loads from wind or accumulation of snow, has several aspects. The maximum load the tubes can support is dependent on their internal pressure; the maximum load the barrier can support further depends on the friction between the extruded tube and the features it was weaved around. 

\begin{figure}[h]
    \centering
    \includegraphics[width =0.4\textwidth]{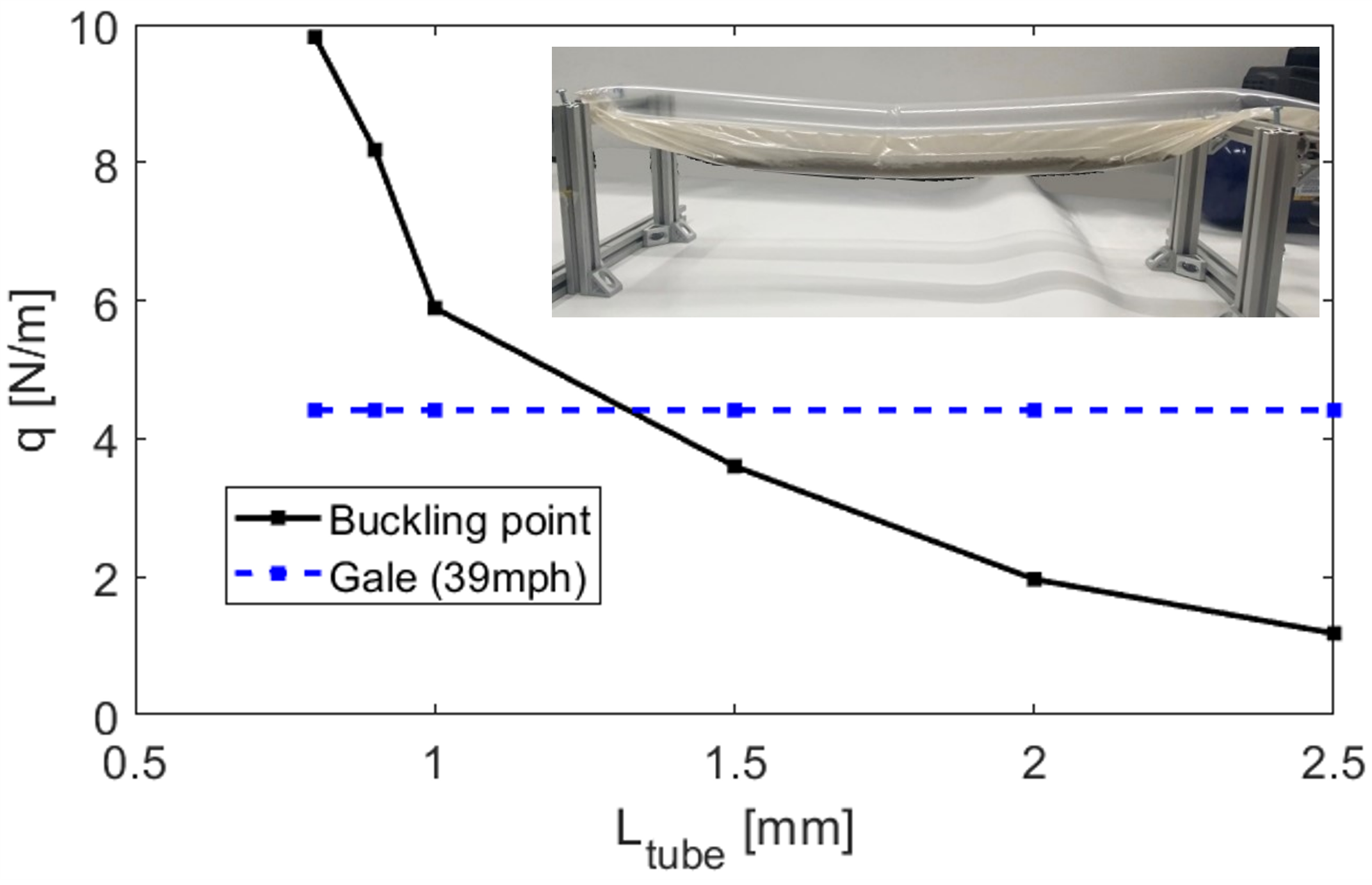}
    \caption{``Buckling point" as a result of distributed load $q$, when the 3" tube with length $L_{tube}$ is pressurized at 8PSI, compared to the resulting load of a gale.
    }
    \label{fig:buckle}
\end{figure}

To gain a general understanding of how the inflated tube would respond to a distributed load under pressure similar to the one we used for inflation, we used the test setup shown in the insert in Fig.~\ref{fig:buckle}. The tube was inflated to 8PSI and suspended between two stationary beams. We attached a second tube that was gradually filled with a high-viscosity material (sand) along its length, and its behavior was observed visually. Qualitatively, we found that under light, distributed loads, the tube acts as a supported cylindrical rod. 
At higher loads, however, the stress causes the tube to kink, concentrating all of the stress in a single location similar to a buckling point. We recorded this ``buckling point" given a distributed load, $q$, as a function of the tube length, $L_{tube}$. 
(Because we used sand to apply the distributed load, the load remains roughly distributed even after the tube is kinked).
The results indicate an exponential correlation between the ``buckling point" and the tube length (Fig.~\ref{fig:buckle}). To ground these results, we compare them to a load from a strong wind ($=1/2 \rho v^2$, where the density of air is $\rho = 1.225 kg/m^3$ and the wind speed in a gale is $v=39mph$) shown in the blue curve. 
These back of the envelope calculations, in addition to traditional bending beam analysis, may help inform future path-planning algorithms to determine which features can be incorporated into the structure.







Future work may also look at optimizing the extrusion mechanism and the path planning toward tighter, more structurally stable weaves. Traditionally woven structures are created by interlocking yarn-like materials in a vertical (warp) and a horizontal (weft) direction. Weft yarns are woven over and under warp yarns in specific sequences to form the surface, and the repeated intersections between the yarns create a frictional force that holds the structure in place~\cite{badawi2023scientific}. On a similar note, the permeability can be tuned by changing the specific sequence of how often the tubes weave over and under the vertical elements in an environment (even double- to triple-layered woven structures could be implemented by varying the vertical elements \cite{sun2020weaving}).  

Finally, it is worth noting that the tube material itself is inextensible and, therefore, strong in tension. Future robots could potentially extrude the tube; use a tip-localized mechanism, akin to a hook, to attach the end of the tube to the environmental features; then deflate and strain the tube, leaving just the material in place. 

\section{Mechanism Demonstration}
\label{sec:demo}

\begin{figure}[t]
    \centering
    \includegraphics[width =0.45\textwidth]{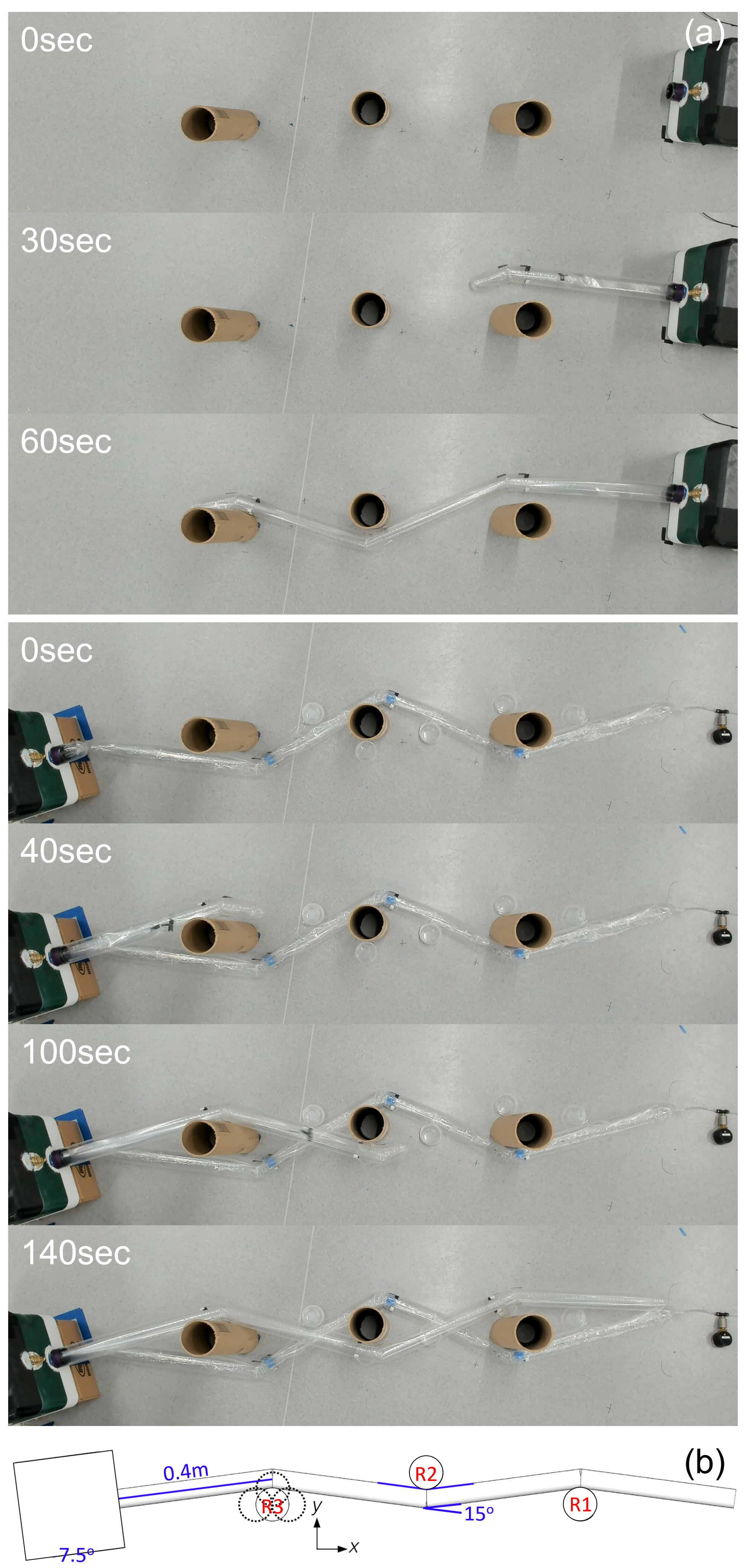}
    \caption{a) Snapshots from the demonstration of two weaved, inflated barrier walls. b) Pillar R3 was moved to examine mechanism resilience to map errors.}
    \label{fig:demo}
\end{figure}

To demonstrate the overall system concept, we placed the extrusion/steering mechanism in front of three pipes. The position of the pipes informed which folds to release, and we executed this sequence on two opposing, layered weaves (Fig.~\ref{fig:demo}(a)). Note that the second tube extruded slower than the first, because the compressor had exhausted much of its air supply. Furthermore, in the demonstration, we extruded tubes from either end of the structure, but these might as well have been extruded from the same side. 

Beyond serving as a proof of concept, this demonstration elucidated two issues that would need to be further addressed in a final system. 
1) Vertical guidance for extrusion of barriers in 3D is necessary. Obviously, to produce multiple tube layers, a robot would either need to lift the entire extrusion mechanism or, perhaps more usefully, have the ability to angle extrusion upwards as well. We found that the second tube sagged under its own weight. In our demonstration, we fixed this by adding support points (plastic cups) in select locations under the tube. In a future robot, extruding at a slight upwards angle could counteract this drooping effect. 
2) Empirically, we found that large bends and large cumulative bend angles produced friction in the everted tube, eventually inhibiting extrusion. This is another factor that future path planners could take into consideration.

Finally, as soft robots are often commended for their ability to handle low-resolution targets, we investigated the effect of limited map accuracy. Specifically, we changed the position of R3 in $\{x,y\}$ to see if the tube could still weave around R2 and R1 (Fig.~\ref{fig:demo}(b)). We found that the mechanism was robust to inaccurate placement of R3 by up to 60mm in the positive $x$ and $y$ directions, but very sensitive in the negative $x$ direction. Even 2-3 millimeters would cause the tube to kink by the outlet, causing it to miss R2. These results merit further investigation in the future, but such risks could be mitigated by planning conservatively around obstacles.    

\section{Path Planning}
\label{sec:alg}


Next, we give a basic algorithm to generate extrusion paths given a map of known features or `objects'. Our reasoning is that plans can be produced ahead of time and sorted based on on-site factors such as the direction of the environmental load, motion restrictions, available material, etc. 
Our approach relies on a modified roadmap path planning algorithm that utilizes spatial decomposition in visibility graphs~\cite{janet1995essential}. The latter suits our system because it produces the shortest path between the desired start and end points, while avoiding object interiors, by using line-of-sight segments similar to the straight-line segments in our tubes. 

\begin{figure}[!b]
\centering
\includegraphics[width=0.4\textwidth]{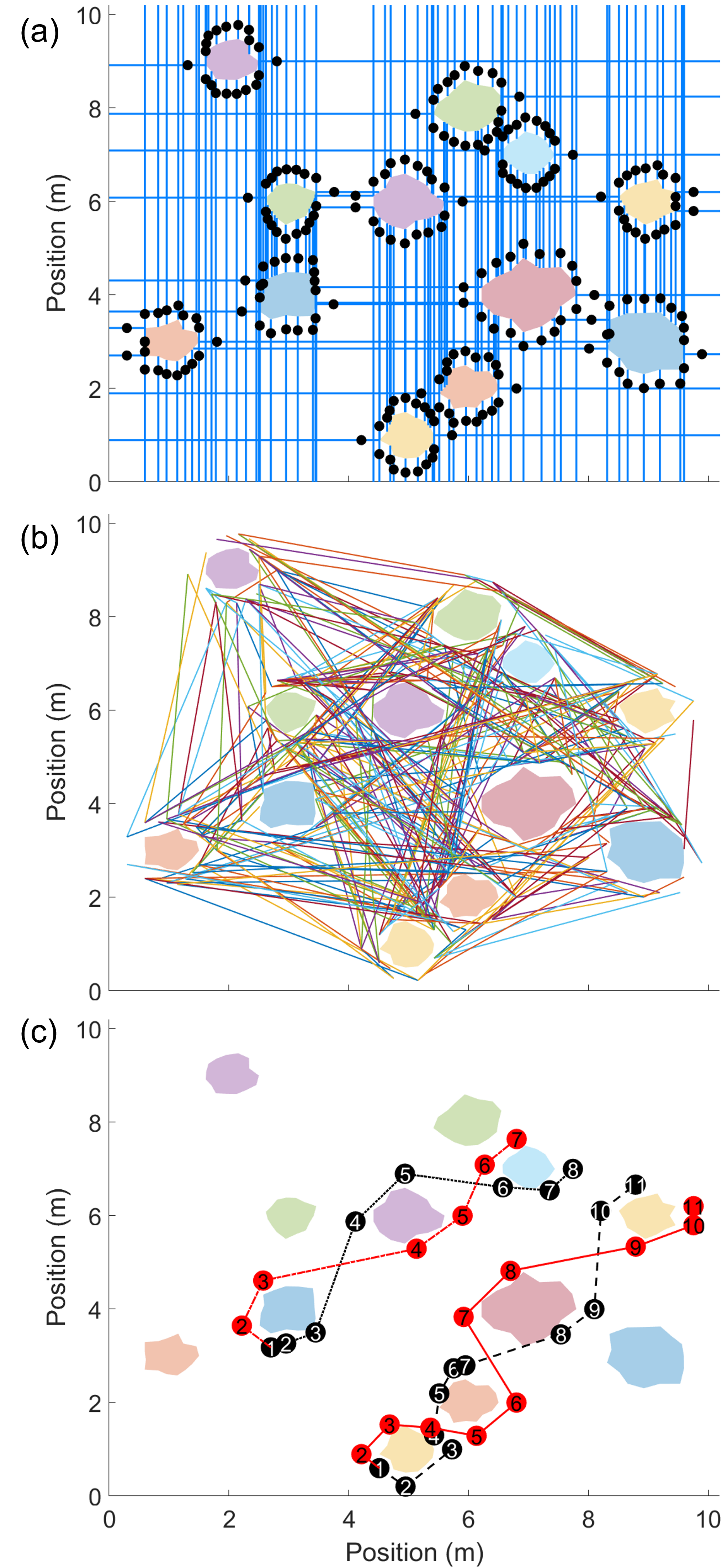}
\caption{(a) Randomized environment with a spatial decomposition (blue lines) generated through visibility graphs and nodes (black dots) in the roadmap. (b) 10\% of paths generated in the visibility map. (c) Example paths formed by the constructed roadmap: counterclockwise-weaved paths shown in black, and clockwise-weaved paths shown in red, with sequentially numbered waypoints.}
\label{fig:path-planning}
\end{figure}

We begin with a set of objects, $\mathcal{O}$. In the simulated example (Fig.~\ref{fig:path-planning}(a)), 12 objects of random shapes and positions were created using code from~\cite{ceron2019comparative}. 
For each object $\mathcal{O}_{i}$, we obtain the set of vertices $\mathcal{V}_{i}$. For each vertex $v$ in $\mathcal{V}_{i}$, we add a node $\mathcal{N}$ to our visibility graph $\mathcal{G}$ by a fixed offset $d$ from $v$ (Fig.~\ref{fig:path-planning}(a) black dots). This procedure creates nodes that surrounds all objects with an offset $d$. 
We then create a roadmap in the form of an adjacency matrix which defines all possible edges between the nodes of $\mathcal{G}$, by finding connections between nodes that do not intersect objects or the environment boundary (Fig.~\ref{fig:path-planning}(b)).

Next, we simply search this graph to generate paths for all pairs of nodes. Examples of two sets of generated weaves can be seen in Fig.~\ref{fig:path-planning}(c). To enforce weaves, we find a subset of waypoints above and below the line that connects the start and end points, and mark these sequentially as clockwise (CW) and counterclockwise (CCW) paths. 
Next, we plan through these waypoints by choosing points from the roadmap in accordance with the current direction (CCW or CW). To obtain the final path, we use a short-range, shortest path algorithm that optimizes over a sliding window of 3 nodes to eliminate unnecessary turns that risk occurring due to a high density of nodes. 

The generated plans lend themselves directly to implementation with the active steering mechanisms mentioned in Sec.~\ref{sec:old_steering}, but would need an additional step to work with the new steering mechanisms proposed in Sec.~\ref{sec:new_steering}. Specifically, we would have to discretize the angles and segment lengths suggested in the path to match with the existing fold frequency, then rerun the simulation to verify that these were still applicable. Related techniques would include 1) dilating object perimeters and representing them with polygons that have fewer vertices, which would produce larger, but fewer bends in the path, and/or 2) increasing the offset, such that inaccurate bend angles or positions were less likely to result in object collisions. 

In a future implementation, it would be useful to eliminate all paths found based on system limitations, i.e. 1) terrain difficulty and navigability constraints, 2) the ability of the tube given its length and internal pressure to withstand the environmental load, and 3) the maximum cumulative bending angle of the path, i.e. the internal friction that the extrusion mechanism must be able to overcome. Upon deployment, remaining paths could then be further sorted based on the desired position and orientation of the barrier and/or environmental loads that the barrier must resist.

\section{DISCUSSION}
\label{sec:conclusion}

In this paper, we presented the initial investigation of the use of everting vine-robots to produce light-duty barriers. In this context, we discussed the benefits of extruded structures composed of lightweight, low-cost materials and the appropriateness of existing steering mechanisms. We further introduced and characterized a novel steering mechanism  which lends itself well to this type of application where tubes are composed of mostly straight lines and intermittent bends in alternating directions. We elaborated on the use of inflated tubes for barriers, the resilience of the mechanism to map errors, and presented a demonstration of two extruded tubes woven around three pillars. Finally, we motivated the use of visibility graphs for path planning given knowledge of existing environmental maps. 

There are many ways to further this work. We are especially excited to pursue the benefits that come from weaving \cite{sun2020weaving}, for example, extrusion patterns that permit tighter weaves and better use of friction between tubes and between tubes and the environmental features. Additionally, combining this mechanism with a real off-terrain robot and exploring the metrics that come from more specific applications, such as human shelters or more complicated geometries, is a natural point of extension.





\bibliographystyle{IEEEtran}
\bibliography{HaronRefs,KirstinRefs}

\end{document}